\documentclass[final]{cvpr}

\usepackage{times}
\usepackage{epsfig}
\usepackage{graphicx}
\usepackage{amsmath}
\usepackage{amssymb}
\usepackage{subfig}
\usepackage{bm}
\usepackage{multirow}  
\usepackage{booktabs}
\usepackage{comment}

\usepackage[table]{xcolor}
\definecolor{mygray}{gray}{.92}

\makeatletter
\newcommand{\thickhline}{%
    \noalign {\ifnum 0=`}\fi \hrule height 1pt
    \futurelet \reserved@a \@xhline
}

\usepackage[pagebackref=true,breaklinks=true,colorlinks,bookmarks=false]{hyperref}

\def\cvprPaperID{2426} 
\def\confYear{CVPR 2021}

\begin{document}

\title{Anchor-Free Person Search}

\author{Yichao Yan$^{1}$ \thanks{indicates equal contributions; ${}^{\dagger}$indicates corresponding authors} $\ ^{\dagger}$ ,
Jinpeng Li$^{1} \footnotemark[1]$ ,
Jie Qin$^{1} {}^{\dagger}$,
Song Bai$^{2}$,
Shengcai Liao$^{1}$,
Li Liu$^{1}$,
Fan Zhu$^{1}$,
and Ling Shao$^{1}$
\\
\noindent
$^{1}$
Inception Institute of Artificial Intelligence (IIAI), UAE \qquad $^{2}$ University of Oxford, UK \qquad \\

$^{}$ {\tt\small \{yanyichao91, ljpadam, qinjiebuaa, songbai.site\}@gmail.com,} \tt\small \{scliao, ling.shao\}@ieee.org\\
}

\maketitle

\begin{abstract}

   Person search aims to simultaneously localize and identify a query person from realistic, uncropped images, which can be regarded as the unified task of pedestrian detection and person re-identification (re-id). Most existing works employ two-stage detectors like Faster-RCNN, yielding encouraging accuracy but with high computational overhead. In this work, we present the Feature-Aligned Person Search Network (AlignPS), \textbf{the first anchor-free framework} to efficiently tackle this challenging task. AlignPS explicitly addresses the major challenges, which we summarize as the misalignment issues in different levels (i.e., scale, region, and task), when accommodating an anchor-free detector for this task.
   More specifically, we propose an aligned feature aggregation module to generate more discriminative and robust feature embeddings by following a ``re-id first" principle. Such a simple design directly improves the baseline anchor-free model on CUHK-SYSU by more than 20\% in mAP. Moreover, AlignPS outperforms state-of-the-art two-stage methods, with a higher speed. Code is available at: {\tt\small \url{ https://github.com/daodaofr/AlignPS }}

\end{abstract}

\section{Introduction}


Person search~\cite{DBLP:conf/cvpr/ZhengZSCYT17,DBLP:conf/cvpr/XiaoLWLW17}, which aims to localize and identify a target person from a gallery of realistic, uncropped scene images, has recently emerged as a practical task with real-world applications. To tackle this task, we need to address two fundamental tasks in computer vision, \ie, pedestrian detection~\cite{DBLP:conf/iccv/OuyangW13,DBLP:conf/cvpr/ZhangBS17} and person re-identification (re-id)~\cite{DBLP:conf/cvpr/FarenzenaBPMC10,DBLP:conf/cvpr/AhmedJM15}. Both detection and re-id are very challenging tasks and have received tremendous attention in the past decade. In person search, we need to not only address the challenges (\eg, occlusions, pose/viewpoint variations, and background clutter) of the two individual tasks, but also pursue a unified and optimized framework to simultaneously perform detection and re-id.

\begin{figure}[t]
\setlength{\abovecaptionskip}{1mm}
\subfloat[Two-step person search framework\label{subfig-1-1}]{%
   \includegraphics[width=\linewidth]{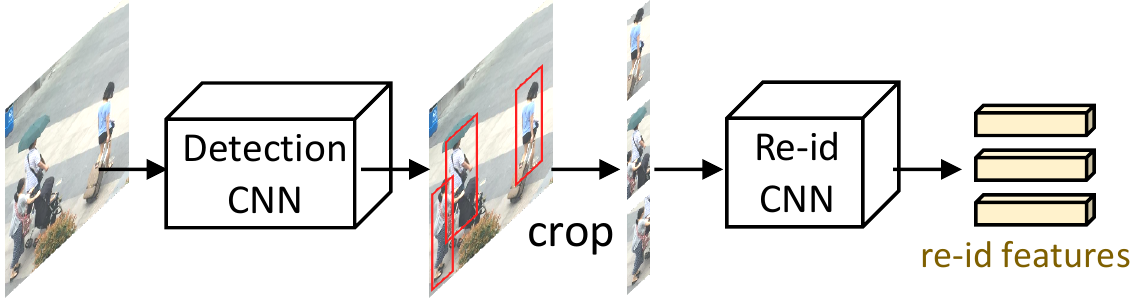}
}
 \vspace{-2mm}
\subfloat[One-step two-stage person search framework\label{subfig-1-2}]{%
   \includegraphics[width=\linewidth]{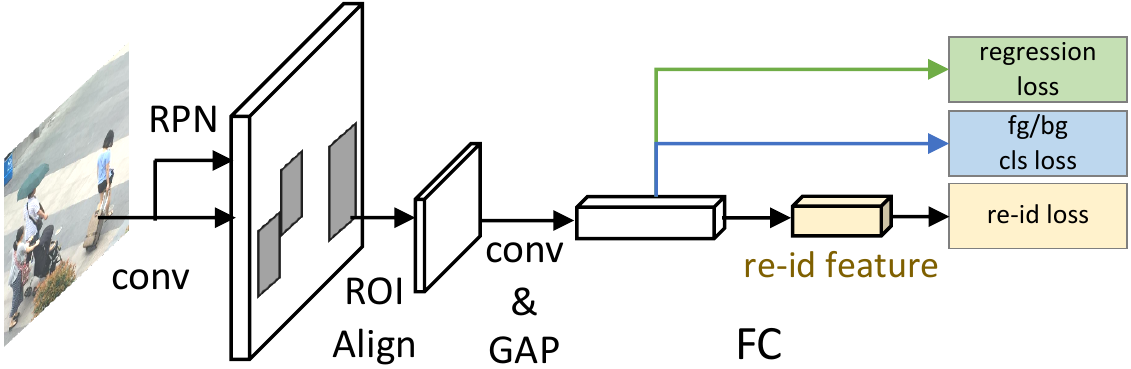}
}
 \vspace{-2mm}
\subfloat[The proposed one-step one-stage anchor-free framework\label{subfig-1-3}]{%
   \includegraphics[width=\linewidth]{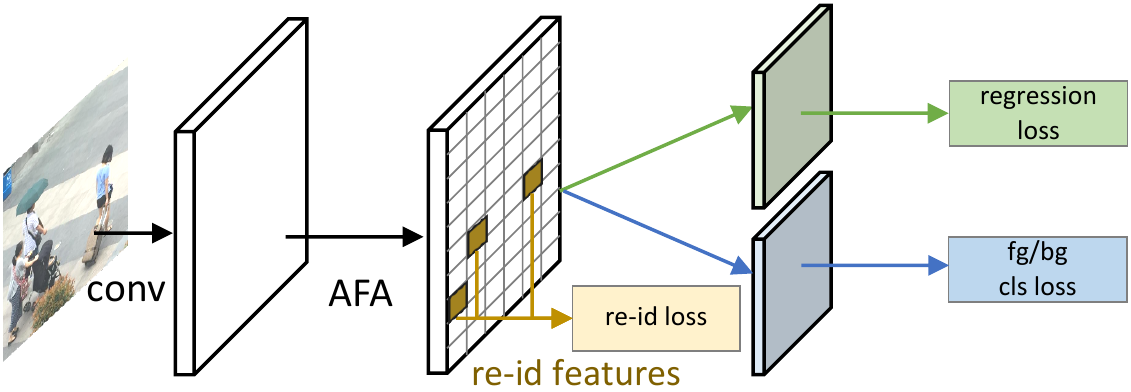}
}
 \caption{Comparison of three person search frameworks. (a) The two-step framework addresses detection and re-id as two separate tasks. (b) The one-step model enables end-to-end training of detection and re-id with an ROI-Align operation based on a two-stage detector; however, re-id is considered as a secondary task after detection. (c) The proposed framework enables single-stage inference for both detection and re-id, while making re-id the primary task.}
 \label{fig:intro}
 \vspace{-6mm}
\end{figure}

Previous efforts devoted to this research topic can be generally divided into two categories. The \emph{first} line of works~\cite{DBLP:conf/cvpr/ZhengZSCYT17,DBLP:conf/eccv/ChenZOYT18,DBLP:conf/eccv/LanZG18}, which we refer to as \emph{two-step} approaches, attempt to deal with detection and re-id separately. As shown in Fig.~\ref{subfig-1-1}, multiple persons are first localized with off-the-shelf detection models, and then cropped out and fed to re-id networks to extract discriminative embeddings. Although two-step models can obtain satisfactory results, the disentangled treatment of the two tasks is time- and resource-consuming.
In contrast, the \emph{second} line of approaches~\cite{DBLP:conf/cvpr/XiaoLWLW17,DBLP:conf/iccv/LiuFJKZQJY17,DBLP:conf/eccv/ChangHSLYH18,DBLP:conf/cvpr/MunjalATG19,DBLP:conf/cvpr/ChenZYS20} provide a \emph{one-step} solution that unifies detection and re-id in an end-to-end manner.
As shown in Fig.~\ref{subfig-1-2}, one-step models first apply an ROI-Align layer to aggregate features in the detected bounding boxes. The features are then shared by detection and re-id; with an additional re-id loss, the simultaneous optimization of the two tasks becomes feasible. Since these models adopt two-stage detectors like Faster-RCNN~\cite{DBLP:journals/pami/RenHG017}, we refer to them as \emph{one-step two-stage} models. However, these methods inevitably inherit the limitations of two-stage detectors, \eg, high computational complexity caused by dense anchors, and high sensitivity to the hyperparameters including the size, aspect ratio and number of anchor boxes, \etc.

In contrast to two-stage detectors, anchor-free models exhibit  unique advantages (\eg, simpler structure and higher speed), and have been actively studied in recent years \cite{DBLP:conf/cvpr/RedmonDGF16,DBLP:conf/eccv/LawD18,DBLP:conf/cvpr/LiuLRHY19,DBLP:conf/iccv/DuanBXQH019}. Inspired by this, an open question is naturally thrown at us - \emph{Is it possible to develop an anchor-free framework for person search?} The answer is yes. However, this is a non-trivial task due to the following three misalignment issues. \textbf{1)} Many anchor-free models learn multi-scale features using feature pyramid networks (FPNs) \cite{DBLP:conf/cvpr/LinDGHHB17} to achieve scale invariance for object detection. However, this introduces the misalignment issue for re-id (\ie, scale misalignment), as a query person needs to be compared with all the people of various scales in the gallery set. \textbf{2)} In the absence of operations like ROI-Align, anchor-free models cannot align the features for re-id and detection according to a specific region. Therefore, re-id embeddings  must be directly learned from feature maps without explicit region alignment. \textbf{3)} Person search can be intuitively formulated as a multi-task learning framework with detection and re-id as its sub-tasks. Hence, we need to find a better tradeoff/alignment between the two tasks.

In this work, we present the first anchor-free framework for efficient person search, which we name the Feature-Aligned Person Search Network (\textbf{AlignPS}). Our model employs the typical architecture of anchor-free detection models, but with a carefully designed aligned feature aggregation (AFA) module. We follow a ``re-id first" principle to explicitly address the above-mentioned challenges. More specifically, AFA reshapes some building blocks of FPN by exploiting the deformable convolution and feature fusion to overcome the issues of region and scale misalignment in re-id feature learning. We also optimize the training procedures of re-id and detection to place more emphasis on generating robust re-id embeddings (as shown in Fig.~\ref{subfig-1-3}). These simple yet effective designs successfully transform a classic anchor-free detector into a powerful and efficient person search framework, and allow the proposed model to outperform its anchor-based competitors.

In summary, our main contributions include:
\begin{itemize}
\setlength{\itemsep}{0pt}
	\setlength{\parsep}{-2pt}
	\setlength{\parskip}{-0pt}
	\setlength{\leftmargin}{-15pt}
	\vspace{-7pt}
    \item We propose the first \emph{one-step} \emph{one-stage} framework for efficient person search. The \emph{anchor-free} solution will significantly foster future research in this direction.
    \item We design an AFA module that simultaneously addresses the issues of scale, region, and task misalignment to successfully accommodate an anchor-free detector for the task of person search.
    \item As an anchor-free one-stage framework, our model surprisingly outperforms state-of-the-art one-step two-stage models on two challenging person search benchmarks, while running at a higher speed.
\end{itemize}

\section{Related Work}
\textbf{Pedestrian Detection}. Pedestrian or object detection can be considered as a preliminary task of person search. Current deep learning-based detectors are generally categorized into one-stage and two-stage models, according to whether they employ a region proposal layer to generate object proposals. Alternatively, object detectors can also be categorized into anchor-based and anchor-free detectors, depending on whether they utilize anchor boxes to associate objects. One of the most representative two-stage anchor-based detectors is Faster-RCNN~\cite{DBLP:journals/pami/RenHG017}, which has been extended into numerous variants~\cite{DBLP:conf/iccv/DaiQXLZHW17,DBLP:conf/cvpr/CaiV18,DBLP:conf/cvpr/PangCSFOL19,DBLP:conf/cvpr/SongLW20}. Notably, some one-stage detectors~\cite{DBLP:conf/eccv/LiuAESRFB16,DBLP:conf/iccv/LinGGHD17,DBLP:conf/cvpr/RedmonF17,DBLP:conf/cvpr/ZhangWBLL18} also work with anchor boxes. Compared with the above models, one-stage anchor-free detectors~\cite{DBLP:conf/cvpr/RedmonDGF16,DBLP:conf/eccv/LawD18,DBLP:conf/cvpr/LiuLRHY19,DBLP:journals/corr/abs-1904-07850,DBLP:conf/iccv/YangLHWL19,DBLP:conf/iccv/TianSCH19} have been attracting more and more attention recently due to their simple structures and 
efficient implementations. In this work, we develop our person search framework based on a classic one-stage anchor-free detector, thus making the whole framework simpler and faster.    

\textbf{Person Re-identification}. Person re-id is also closely related to person search, aiming to learn identity embeddings from cropped person images. Traditional methods employed various handcrafted features~\cite{DBLP:journals/ijcv/Lowe04,DBLP:conf/cvpr/FarenzenaBPMC10,DBLP:conf/eccv/GrayT08} before the renaissance of deep learning. However, to pursue better performance, current re-id models are mostly based on deep learning. Some models employ structure/part information in the human body to learn more robust representations~\cite{DBLP:conf/iccv/SuLZX0T17,DBLP:conf/eccv/SunZYTW18,DBLP:conf/iccv/MiaoWLD019,9233968}, while others focus on learning better distance metrics~\cite{DBLP:conf/cvpr/AhmedJM15,DBLP:journals/corr/HermansBL17,DBLP:conf/cvpr/ChenCZH17,DBLP:journals/pami/ChenZZL18,DBLP:journals/pami/WangGZW16}. As person re-id usually lacks large-scale training data, data augmentation~\cite{DBLP:conf/nips/GeLZYYWL18,DBLP:conf/cvpr/LiuNYZCH18,DBLP:conf/cvpr/WeiZ0018,DBLP:conf/iccv/ZhengZY17} also becomes popular for tackling this task. Compared with detection which aims to learn common features of pedestrians, re-id needs to focus more on fine-grained details and unique features of each identity. Therefore, we propose to follow the ``re-id first" principle to raise the priority of the re-id task, resulting in more discriminative identity embeddings for more accurate person search.

\begin{figure*}[t]
\setlength{\abovecaptionskip}{1mm}
\centering
\includegraphics[width=\linewidth]{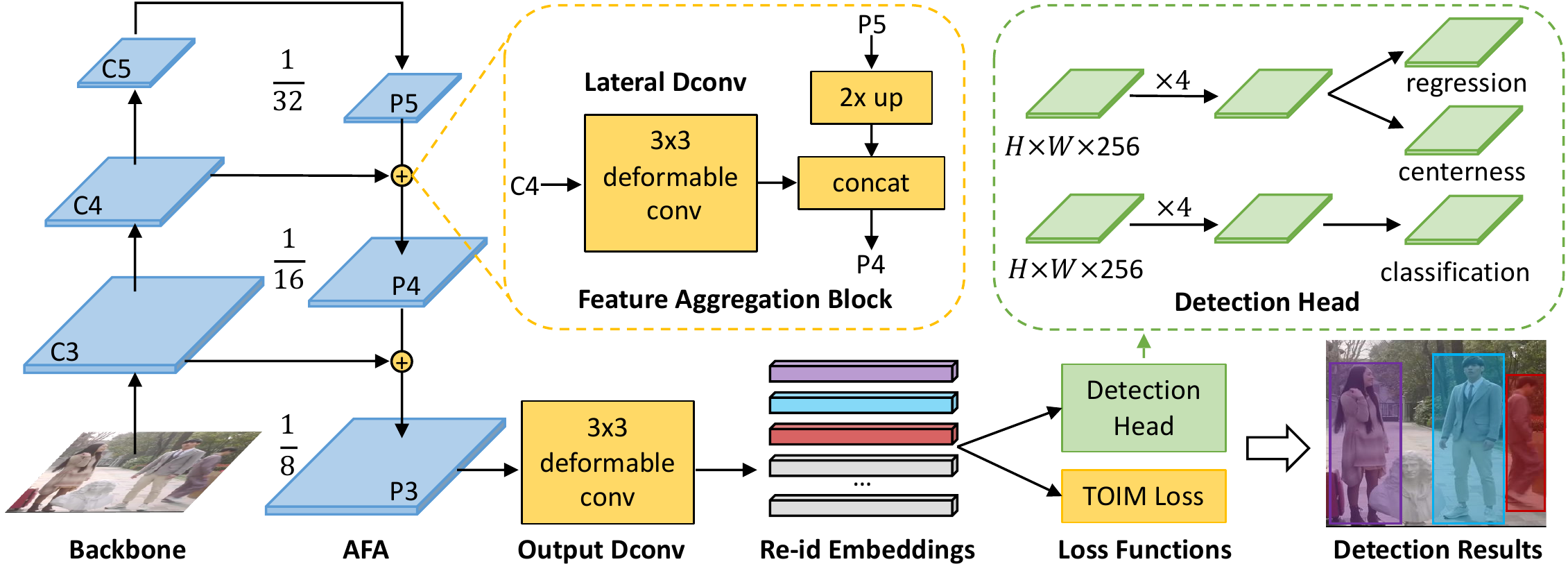}
\caption{Architecture of the proposed AlignPS framework, which shares the basic structure of FCOS~\cite{DBLP:conf/iccv/TianSCH19}. The components in yellow are newly designed to accommodate FCOS for the task of person search. ``Dconv'' means deformable convolution.}
\label{fig:arch}
\vspace{-2mm}
\end{figure*}

\textbf{Person Search}. Existing person search frameworks can be divided into two-step and one-step models. Two-step models first perform pedestrian detection and subsequently crop the detected people for re-id. Zheng \etal~\cite{DBLP:conf/cvpr/ZhengZSCYT17} introduced the first two-step framework for person search and evaluated the combinations of different detectors and re-id models. Since then, several models~\cite{DBLP:conf/eccv/ChenZOYT18,DBLP:conf/eccv/LanZG18,DBLP:conf/iccv/HanYZTZGS19,DBLP:conf/cvpr/WangMCSC20} have followed this pipeline. In~\cite{DBLP:conf/cvpr/XiaoLWLW17}, Xiao \etal proposed the first one-step person search framework based on Faster-RCNN. Specifically, a joint framework enabling end-to-end training of detection and re-id was proposed by stacking a re-id embedding layer after the detection features and proposing the Online Instance Matching (OIM) loss. So far, a number of improvements~\cite{DBLP:conf/iccv/LiuFJKZQJY17,DBLP:journals/pr/XiaoXTHWF19,DBLP:conf/eccv/ChangHSLYH18,DBLP:conf/cvpr/YanZNZXY19,DBLP:conf/cvpr/MunjalATG19,DBLP:conf/cvpr/DongZST20a,DBLP:conf/cvpr/ChenZYS20} have been made based on this framework. In general, two-step models may achieve better performance, while one-step models have the advantages of simplicity and efficiency. However, there is still room for improving one-step methods due to the aforementioned shortcomings of the two-stage anchor-based detectors they usually adopt. In this work, we introduce the first anchor-free model to further improve the simplicity and efficiency of one-step models, without any sacrifice in accuracy. 

\section{Feature-Aligned Person Search Networks}
In this section, we introduce the proposed anchor-free framework (\ie, AlignPS) for person search. Firstly, we give an overview of the network architecture. Secondly, the proposed AFA module is elaborated with the aim of mitigating different levels of misalignment issues when transforming an anchor-free detector into a superior person search framework. Finally, we present the designed loss function to obtain more discriminative features for person search.

\subsection{Framework Overview}
The basic framework of the proposed AlignPS is based on FCOS~\cite{DBLP:conf/iccv/TianSCH19}, one of the most popular one-stage anchor-free object detectors. Differently, we adhere to the ``re-id first" principle to put emphasis on learning robust feature embeddings for the re-id subtask, which is crucial for enhancing the overall performance of person search.

As illustrated in Fig.~\ref{fig:arch}, our model simultaneously localizes multiple people in the image and learns re-id embeddings for them. Specifically, an AFA module is developed to aggregate features from multi-level feature maps in the backbone network. To learn re-id embeddings, which is the key of our method, we directly take the flattened features from the output feature maps of AFA as the final embeddings, without any extra embedding layers. For detection, we employ the detection head from FCOS which is good enough for the detection subtask. The detection head consists of two branches, both of which contain four 3$\times$3 \emph{conv} layers. In the meantime, the first branch predicts regression offsets and centerness scores, while the second makes foreground/background classification. Finally, each location on the output feature map of AFA will be associated with a bounding box with classification and centerness scores, as well as a re-id feature embedding.

\subsection{Aligned Feature Aggregation}
Following FPN~\cite{DBLP:conf/cvpr/LinDGHHB17}, we make use of different levels of feature maps to learn detection and re-id features. As the key of our framework, the proposed AFA performs three levels of alignment, beyond the original FPN, to make the output re-id features more discriminative.

\textbf{Scale Alignment}. The original FCOS model employs different levels of features to detect objects of different sizes. This significantly improves the detection performance since the overlapped ambiguous samples will be assigned to different layers. For the re-id task, however, the multi-level prediction could cause feature misalignment between different scales.
In other words, when matching a person of different scales, re-id features are inconsistently taken from different levels of FPN.
Furthermore, the people in the gallery set are of various scales, which could eventually make the multi-level model fail to find correct matches for the query person. Therefore, in our framework, we only make predictions based on a single layer of AFA, which explicitly addresses the feature misalignment caused by scale variations. Specifically, we employ the $\{C_3, C_4, C_5\}$ feature maps from the ResNet-50 backbone, and AFA sequentially outputs $\{P_5, P_4, P_3\}$, with strides of 32, 16, and 8, respectively. We only learn features from $\{P_3\}$, which is the largest output feature map, for both the detection and re-id subtasks, and $\{P_6, P_7\}$ are no longer generated as in the original FPN. Although this design may slightly influence the detection performance, we will show in Sec.~\ref{sec:analytical} that it achieves a good trade-off between the detection and re-id subtasks.

\textbf{Region Alignment}. On the output feature map of AFA, each location perceives the information from the whole input image based on a large receptive field. Due to the lack of the ROI-Align operation as in Faster-RCNN, it is difficult for our anchor-free framework to learn more accurate features within the pedestrian bounding boxes, and thus leading to the issue of region misalignment. The re-id subtask is even more sensitive to this issue as background features could greatly impact the discriminative capability of the learned features. In AlignPS, we address this issue from three perspectives. \emph{First}, we replace the 1$\times$1 \emph{conv} layers in the lateral connections with 3$\times$3 \emph{deformable conv} layers. As the original lateral connections are designed to reduce the channels of feature maps, a 1$\times$1 \emph{conv} is enough. In our design, moreover, the 3$\times$3 \emph{deformable conv} enables the network to adaptively adjust the receptive field on the input feature maps, thus implicitly fulfilling region alignment. \emph{Second}, we replace the ``sum" operation in the top-down pathway with a ``concatenation" operation, which can better aggregate multi-level features. \emph{Third}, we again replace the 3$\times$3 \emph{conv} with a 3$\times$3 \emph{deformable conv} for the output layer of FPN, which further aligns the multi-level features to finally generate a more accurate feature map. The above three designs work seamlessly to address the region misalignment issue, and we notice that these simple designs are extremely effective when accommodating the basic anchor-free model for our person search task.

\textbf{Task Alignment}. Existing person search frameworks typically treat pedestrian detection as the primary task, \ie, re-id embeddings are just generated by stacking an additional layer after the detection features. A recent work~\cite{DBLP:journals/corr/abs-2004-01888} investigated a parallel structure by employing independent heads for the two tasks to achieve robust multiple object tracking results. In our task of person search, we find the inferior re-id features largely hinder the overall performance. Therefore, we opt for a different principle to align these two tasks by treating re-id as our primary task. Specifically, the output features of AFA are directly supervised with a re-id loss (which will be introduced in the following subsection), and then fed to the detection head. This ``re-id first" design is based on two considerations. \emph{First}, the detection subtask has been relatively well addressed by existing person search frameworks, which directly inherit the advantages from existing powerful detection frameworks. Therefore, learning discriminative re-id embeddings is our primary concern. As we discussed, re-id performance is more sensitive to region misalignment in an anchor-free framework. Therefore, 
it is desirable for the person search framework to be inclined towards the re-id subtask. We also show in our experiments that this design significantly improves the discriminative capability of the re-id embeddings, while having negligible impact on detection. \emph{Second}, compared with ``detection first" and parallel structures, the proposed ``re-id first" structure does not require an extra layer to generate re-id embeddings, and is thus more efficient.

\begin{figure}[t]
\setlength{\abovecaptionskip}{1mm}
\centering
\includegraphics[width=\linewidth]{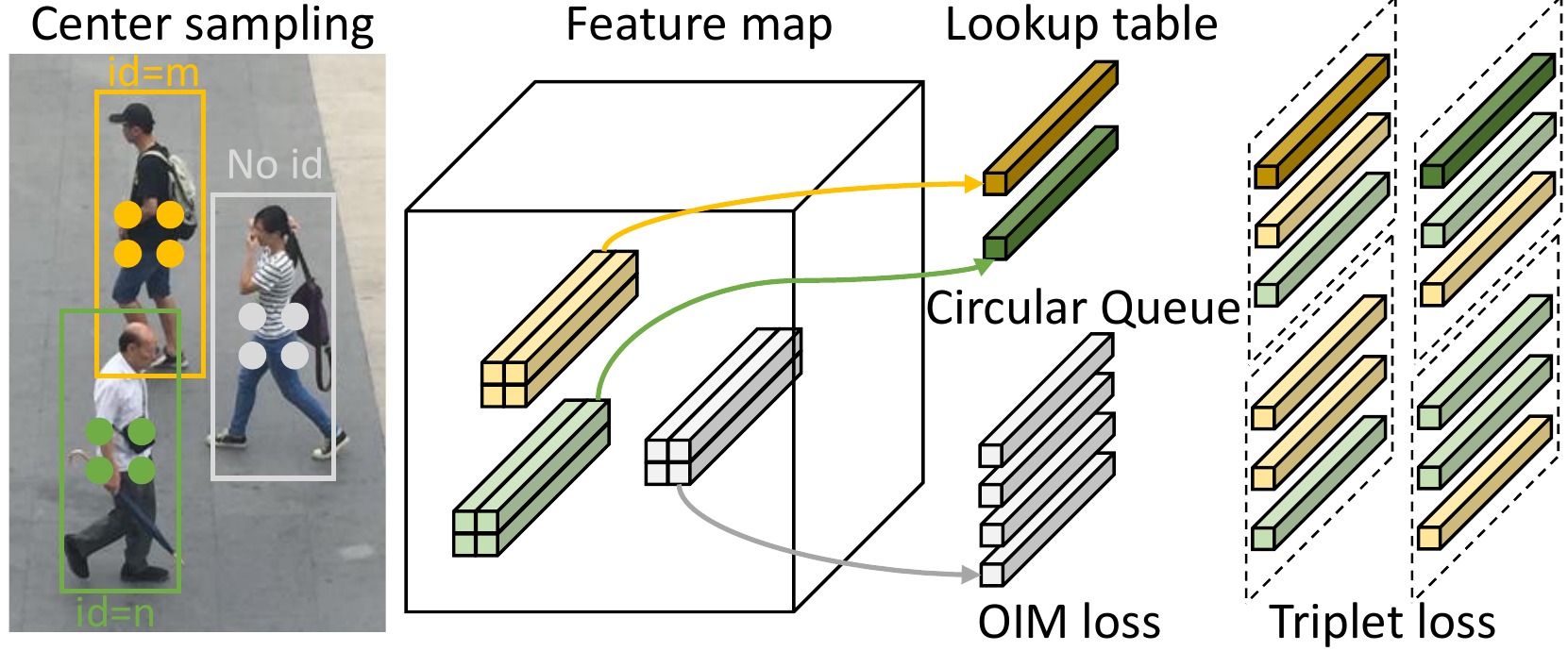}
\caption{Illustration of the Triplet-aided Online Instance Matching loss, where both the features from the input image and the lookup table are sampled to form the triplet.}
\label{fig:loss}
\end{figure}

\subsection{Triplet-Aided Online Instance Matching Loss}
Existing works typically employ the OIM loss to supervise the training of the re-id subtask. Specifically, OIM stores the feature centers of all labeled identities in a lookup table (LUT), $V \in \mathbb{R}^{D \times L} = \{v_1,...,v_L\}$, which contains $L$ feature vectors with $D$ dimensions. Meanwhile, a circular queue $U \in \mathbb{R}^{D \times Q}= \{u_1,...,u_Q\}$ containing the features of $Q$ unlabeled identities is maintained. At each iteration, given an input feature $x$ with label $i$, OIM computes the similarity between $x$ and all the features in the LUT and circular queue by $V^T x$ and $Q^T x$, respectively. The probability of $x$ belonging to the identity $i$ is calculated as:
\begin{equation}
    p_i = \frac{{\rm exp}(v_i^T x) / \tau}{\sum_{j=1}^{L}{\rm exp}(v_j^T x) / \tau + \sum_{k=1}^{Q}{\rm exp}(u_k^T x) / \tau},
\end{equation}
where $\tau=0.1$ is a hyperparameter that controls the softness of the probability distribution. The objective of OIM is to minimize the expected negative log-likelihood:
\begin{equation}
    \mathcal{L}_{\text{OIM}} = -{\rm E}_x[{\rm log}~p_t], \ t = 1, 2, ..., L.
\end{equation}

Although OIM effectively employs both labeled and unlabeled samples, we still observe two limitations. First, the distances are only computed between the input features and the features stored in the lookup table and circular queue, while no comparisons are made between the input features. Second, the log-likelihood loss term does not give an explicit distance metric between feature pairs. 

To improve OIM, we propose a specifically designed triplet loss. For each person in the input images, we employ the center sampling strategy as in~\cite{DBLP:journals/tip/KongSLJLS20}. As shown in Fig.~\ref{fig:loss}, for each person, a set of features located around the person center are considered as positive samples. The objective is to pull the feature vectors from the same person close, and push the vectors from different people away. Meanwhile, the features from the labeled persons should be close to the corresponding features stored in the LUT, and away from the other features in the LUT.

More specifically, suppose we sample $S$ vectors from one person; we get $X_m = \{x_{m,1},..., x_{m,S}, v_m\}$ and $X_n=\{x_{n,1},..., x_{m,S}, v_n\}$ as the candidate feature sets for the persons with identity labels $m$ and $n$, respectively, where $x_{i,j}$ denotes the $j$-th feature of person $i$, and $v_i$ is the $i$-th feature in the LUT. Given $X_m$ and $X_n$, positive pairs can be sampled within each set, while negative pairs are sampled between the two sets. The triplet loss can be calculated as:
\begin{equation}
    \mathcal{L}_{\text{tri}} = \sum_{\text{pos,~neg}}[M + D_{\text{pos}} - D_{\text{neg}}],
\end{equation}
where $M$ denotes the distance margin, and $D_{\text{pos}}$ and $D_{\text{neg}}$ denote the Euclidean distances between the positive pair and the negative pair, respectively. Finally, the Triplet-aided OIM (TOIM) loss is the summation of these two terms:
\begin{equation}
    \mathcal{L}_{\text{TOIM}} = \mathcal{L}_{\text{tri}}+\mathcal{L}_{\text{OIM}}.
\end{equation}

\section{Experiments}

\subsection{Datasets and Settings}
\textbf{CUHK-SYSU}~\cite{DBLP:conf/cvpr/XiaoLWLW17} is a large-scale person search dataset which contains 18,184 images, with 8,432 different identities and 96,143 annotated bounding boxes. The images come from two kinds of data sources (\ie, real street snaps and movies/TV),
covering diverse scenes and including variations of viewpoints, lighting, resolutions, and occlusions.
We utilize the standard training/test split, where the training set contains 5,532 identities and 11,206 images, and the test set contains 2,900 query persons and 6,978 images. This dataset also defines a set of protocols with gallery sizes ranging from 50 to 4,000. We report the results using the default gallery size of 100 unless otherwise specified.

\textbf{PRW}~\cite{DBLP:conf/cvpr/ZhengZSCYT17} was captured using six static cameras in a university campus.
The images are sampled from the videos, which consist of 11,816 video frames in total. Person identities and bounding boxes are manually annotated, resulting in 932 labeled persons with 43,110 bounding boxes. The dataset is split into a training set of 5,704 images with 482 different identities, and a test set of 2,057 query persons and 6,112 images. 

\textbf{Evaluation Metric}. We employ the mean average precision (mAP) and top-1 accuracy to evaluate the performance for person search. We also employ recall and average precision (AP) to measure the detection performance.

\begin{figure}[t]
\setlength{\abovecaptionskip}{1mm}
\centering
\includegraphics[width=\linewidth]{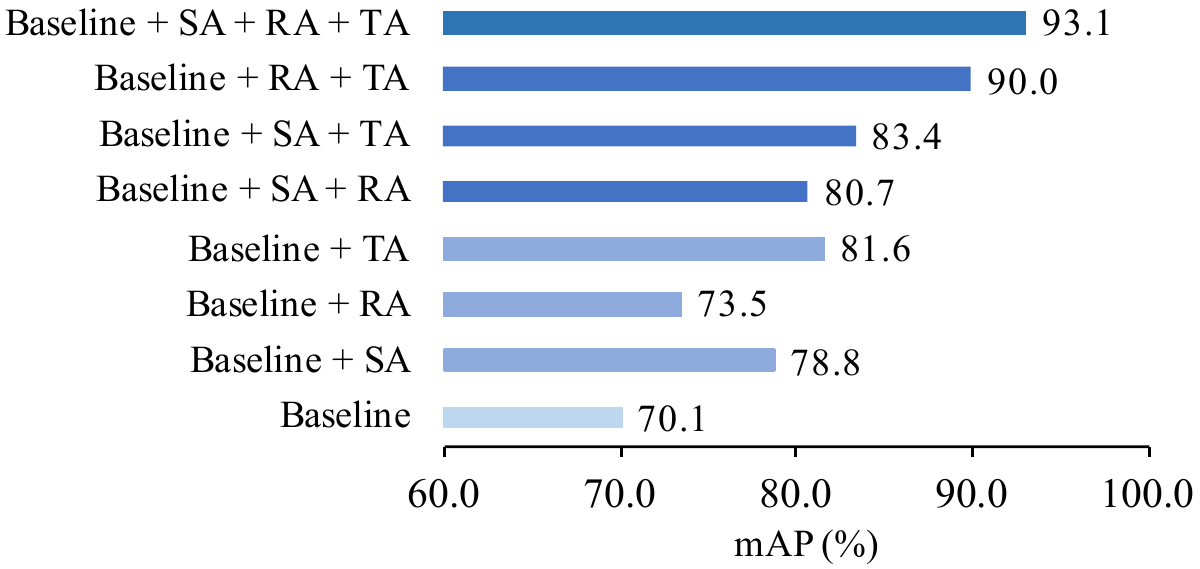}
\caption{Comparative results on CUHK-SYSU with different alignment strategies, \ie, scale alignment (SA), region alignment (RA), and task alignment (TA).}
\label{fig:baseline}
\end{figure}

\subsection{Implementation Details}
We employ ResNet-50~\cite{DBLP:conf/cvpr/HeZRS16} pretrained on ImageNet~\cite{DBLP:conf/cvpr/DengDSLL009} as the backbone. 
We set the batch size to 4, and adopt the stochastic gradient descent (SGD) optimizer with weight decay of 0.0005. The initial learning rate is set to 0.001 and is reduced by a factor of 10 at epoch 16 and 22, with a total of 24 epochs. We use a warmup strategy for 300 steps. We employ a multi-scale training strategy, where the longer side of the image is randomly resized from 667 to 2000 during training, while zero padding is utilized to fit the images with different resolutions. For inference, we rescale the test images to a fixed size of 1500$\times$900. Following~\cite{DBLP:conf/aaai/ChenZO0S20}, we add a focal loss~\cite{DBLP:conf/iccv/LinGGHD17} to the original OIM loss.
All the experiments are implemented based on PyTorch~\cite{DBLP:conf/nips/PaszkeGMLBCKLGA19} and MMDetection~\cite{DBLP:journals/corr/abs-1906-07155}, with an NVIDIA Tesla V100 GPU. It takes around 29 and 20 hours to finish training on CUHK-SYSU and PRW, respectively.

\subsection{Analytical Results}\label{sec:analytical}

\textbf{Baseline}.
We directly add a re-id head in parallel with the detection head to the FCOS model and take it as our baseline. As shown in Fig.~\ref{fig:baseline}, each of the alignment strategies brings notable improvements to the baseline, and combining all of them yields \textgreater20\% improvements in mAP.

\begin{table}[t]
\setlength{\abovecaptionskip}{1mm}
\centering
\begin{tabular}{p{1.7cm}|p{1.1cm}<{\centering}p{1.1cm}<{\centering}|p{1.1cm}<{\centering}p{1.1cm}<{\centering}}
\hline\thickhline
\rowcolor{mygray} 
  & \multicolumn{2}{c|}{Detection} & \multicolumn{2}{c}{Re-id}   \\ \cline{2-5} 
\rowcolor{mygray} 
\multirow{-2}{*}{Methods}  & Recall & AP  & mAP  & top-1  \\  \hline \hline     
$P_3$
  & 90.3    & \textbf{81.2} &\textbf{93.1} & \textbf{93.4}  \\
$P_4$  & 87.5   & 78.7       & 92.7   & 93.1    \\ 
$P_5$  & 79.0   & 71.7       & 89.3   & 89.5  \\
$P_3$, $P_4$  & 90.4  & 80.5       & 91.1 & 91.6 \\ 
$P_3$, $P_4$, $P_5$  & \textbf{90.9}  & 80.4       & 90.0 & 90.5 \\\hline
\end{tabular}
\caption{Comparative results on CUHK-SYSU by employing different levels of features. $P_3$, $P_4$, and $P_5$ are the feature maps with strides of 8, 16, and 32, respectively. }
\label{tab:scale}
\end{table}

\textbf{Scale Alignment}. 
To evaluate the effects of scale alignment, we employ feature maps from different levels of AFA and report the results in Table~\ref{tab:scale}. Specifically, we evaluate the features from $P_3$, $P_4$, and $P_5$ with strides of 8, 16, and 32, respectively. As can be observed, features from the largest scale $P_3$ yield the best performance, due to the fact that they absorb different levels of features from AFA, providing richer information for detection and re-id. Similar to FCOS, we also evaluate the performance by assigning people of different scales to different feature levels. We set the size ranges for \{$P_3, P_4\}$ as [0, 128] and [128, $\infty$], while the prediction ranges for \{$P_3, P_4, P_5\}$ are [0, 128], [128, 256], and [256, $\infty$], respectively. We can see that these dividing strategies achieve slightly better detection results w.r.t. the recall rate. However, they bring back the scale misalignment issue to person re-id. Also note that this issue is not well addressed with the multi-scale training strategy. All the above results demonstrate the necessity and effectiveness of the proposed scale alignment strategy.

\begin{table}[t]
\setlength{\abovecaptionskip}{1mm}
\centering
\begin{tabular}{p{1.2cm}<{\centering}p{1.3cm}<{\centering}p{1.2cm}<{\centering}|p{1.2cm}<{\centering}p{1.2cm}<{\centering}}
\hline\thickhline
\rowcolor{mygray}  
Lateral              & Output               & Feature              & \multicolumn{2}{c}{Re-id}                        \\ \cline{4-5} 
\rowcolor{mygray}  
dconv                 & dconv                 & concat               & mAP                  & \multicolumn{1}{c}{top-1} \\ 
\hline \hline  
 &   & & 83.4 & 83.7                         \\ 
$\surd$ &   & & 90.6 & 90.8                       \\
 & $\surd$  & & 91.4 & 91.9                           \\
  &   & $\surd$ & 84.0 & 84.1                           \\
$\surd$ & $\surd$  & & 91.8 & 92.2                           \\
$\surd$ &   & $\surd$ & 90.7 & 91.0                            \\
 & $\surd$  & $\surd$& 92.0 & 92.5                            \\
$\surd$ & $\surd$  & $\surd$ & \textbf{93.1} & \textbf{93.4}                         \\\hline
\end{tabular}
\caption{Comparative results on CUHK-SYSU by employing different components in AFA for region alignment. ``dconv'' stands for deformable convolution. }
\label{tab:region}
\end{table}

\begin{figure}[]
\setlength{\abovecaptionskip}{1mm}
\begin{center}
\subfloat[Deformable conv at lateral $C_3$ layer in AFA\label{subfig-dcnvis-1}]{%
   \includegraphics[width=\linewidth]{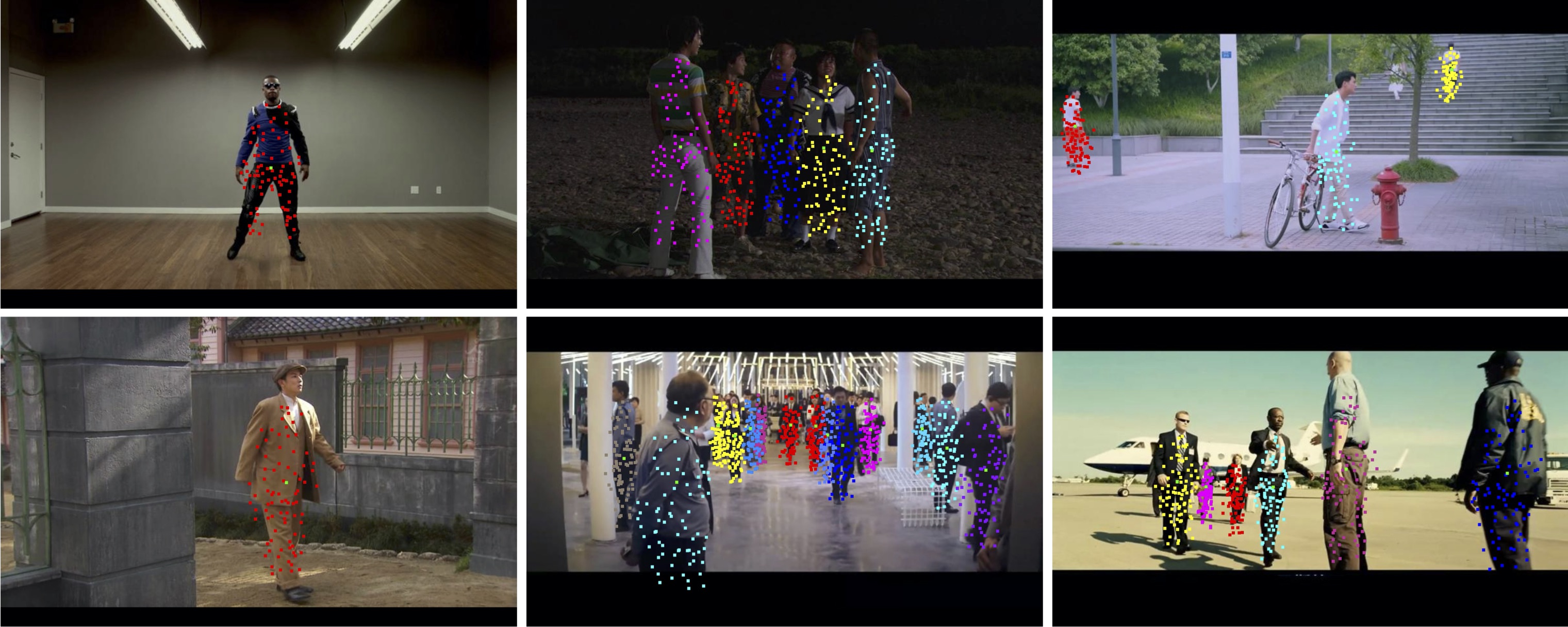}
}
 \hfill
 \vspace{-4mm}
\subfloat[Deformable conv at lateral $C_4$ layer in AFA\label{subfig-dcnvis-2}]{%
   \includegraphics[width=\linewidth]{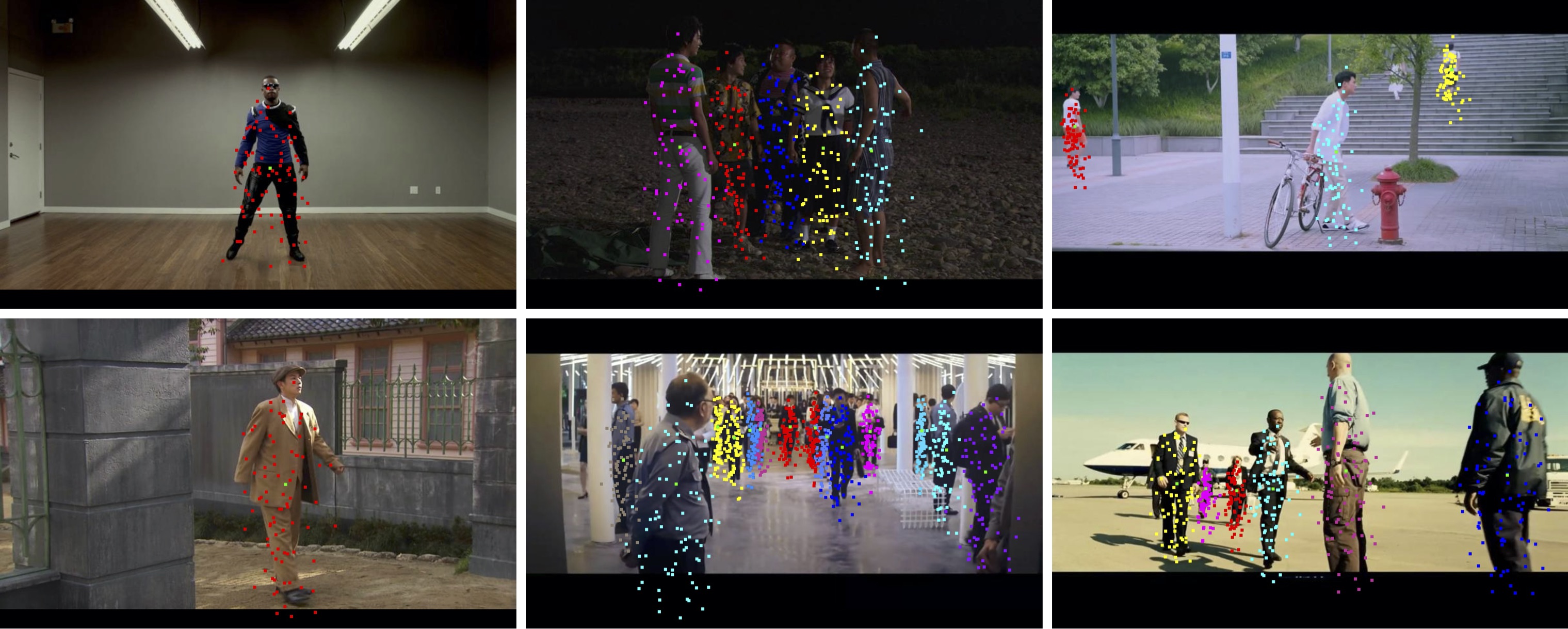}
}
\end{center}
\vspace{-4mm}
   \caption{Each image shows the sampling locations of two levels of 3$\times$3 ($9^2 = 81$ points at each location) deformable filters: (a) Lateral deformable conv $C_3$ + Output deformable conv; (b) Lateral deformable conv $C_4$ + Output deformable conv. We illustrate different locations with different colors, while center locations of people are marked in green. Please zoom in for better visualization.}
\label{fig:dcnvis}
\end{figure}

\textbf{Region Alignment}. 
We conduct experiments with different combinations of lateral deformable conv, output deformable conv and feature concatenation, and analyze how different region alignment components influence the overall performance. The results are reported in Table~\ref{tab:region}. Without all these modules, the framework only achieves 83.7\% in top-1 accuracy, which is $\sim$10\% lower than the full model. The individual components of lateral deformable conv and output deformable conv improve the model by $\sim$7\% and $\sim$8\%, respectively. Feature concatenation also brings $\sim$1\% improvements. By combining two of the three components, we observe consistent improvements. Finally, employing all the three modules yields 93.1\% in mAP and 93.4\% in top-1 accuracy, significantly boosting the performance. These ablation studies thoroughly demonstrate the effectiveness of the region alignment strategies.

To further illustrate how the deformable convolutions work in our framework, we visualize the learned offsets of the deformable filters in Fig.~\ref{fig:dcnvis}. We observe that the proposed framework is capable of learning adaptive receptive field according to the layout of the human body, and is robust to occlusion, crowding, and scale variations. We also observe that the lateral deformable conv in $C_3$ learns tighter offsets around the body center, while the offsets in the $C_4$ layer cover larger regions, which makes the two layers complementary to each other.

\begin{figure}[]
\vspace{-4mm}
\setlength{\abovecaptionskip}{1mm}
\centering
\subfloat[$T_1$\label{subfig-2-1}]{%
   \includegraphics[width=0.485\linewidth]{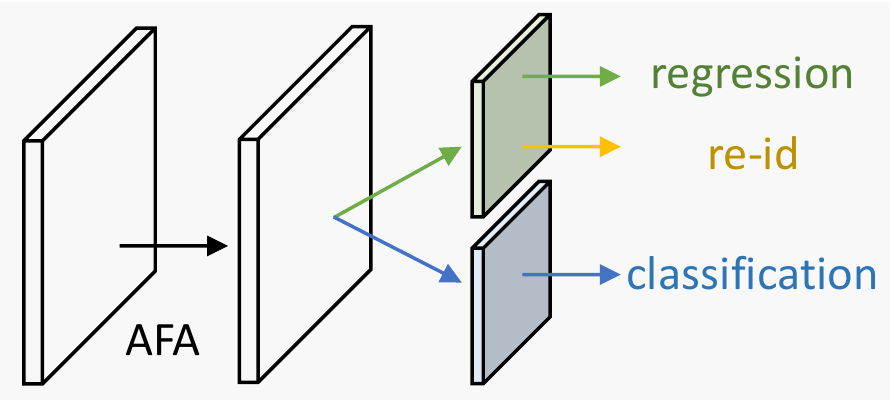}
}
 \hfill
\subfloat[$T_2$\label{subfig-2-2}]{%
   \includegraphics[width=0.485\linewidth]{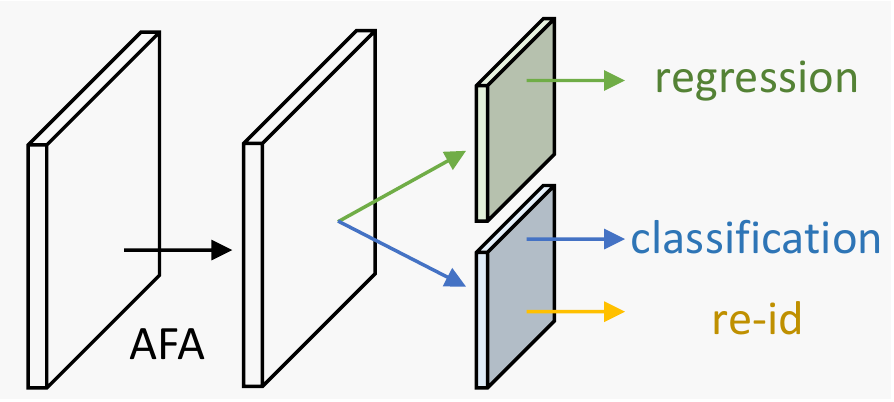}
}
 \hfill
 \vspace{-4mm}
\subfloat[$T_3$\label{subfig-2-3}]{%
   \includegraphics[width=0.485\linewidth]{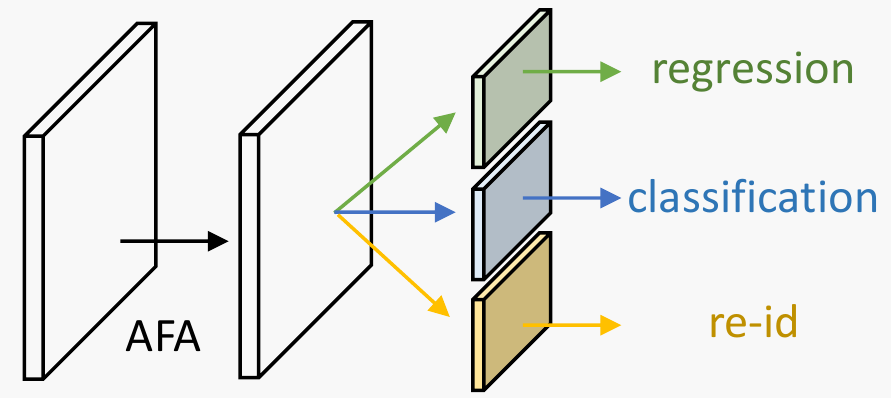}
}
 \hfill
\subfloat[AlignPS\label{subfig-2-4}]{%
   \includegraphics[width=0.485\linewidth]{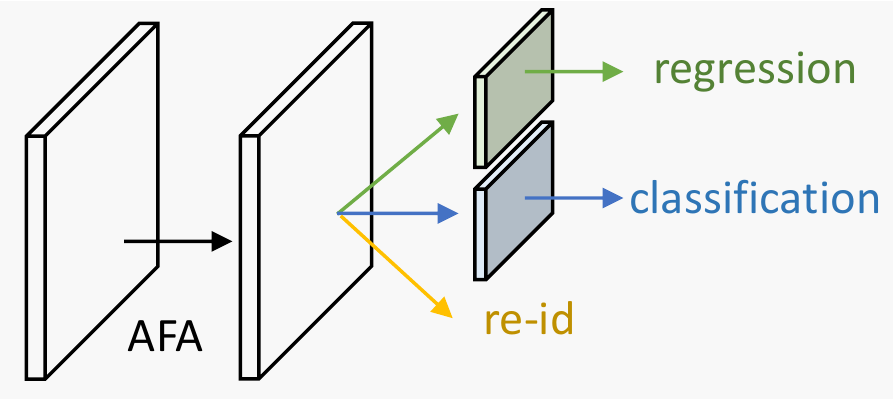}
}
 \caption{Illustration of different structures for training the detection and re-id tasks.}
 \label{fig:tasks}
\end{figure}

\textbf{Task Alignment}.
Since person search aims to simultaneously address detection and re-id subtasks in a single framework, it is important to understand how different configurations of the two subtasks influence the overall task and which subtask should be paid more attention to. To this end, we design several structures to compare different training options (as shown in Fig.~\ref{fig:tasks}), the performance of which is summarized in Table~\ref{tab:tasks}. As can be observed, the structures of $T_1$ and $T_2$, where re-id features are shared with the regression and classification heads, respectively, yield significantly lower performance in re-id compared with our design. This indicates that the detection task takes advantage of the shared heads. As for $T_3$ where re-id and detection have independent feature heads, it achieves slightly better performance compared with $T_1$ and $T_2$, but still remarkably underperforms our design. 
These results indicate that our ``re-id first'' structure achieves the best task alignment among all these designs.

\begin{table}[t]
\setlength{\abovecaptionskip}{1mm}
\centering
\begin{tabular}{p{1.8cm}|p{1.1cm}<{\centering}p{1.1cm}<{\centering}|p{1.1cm}<{\centering}p{1.1cm}<{\centering}}
\hline\thickhline
\rowcolor{mygray} 
  & \multicolumn{2}{c|}{Detection} & \multicolumn{2}{c}{Re-id}   \\ \cline{2-5} 
\rowcolor{mygray} 
\multirow{-2}{*}{Methods}  & Recall & AP  & mAP  & top-1  \\  \hline \hline     
$T_1$
   & 87.5   & 79.0       & 80.3   & 79.2   \\
$T_2$  & 89.1   & 78.6      & 77.1   & 75.9    \\ 
$T_3$  & 90.1   & 81.4       & 80.7   & 80.2  \\
AlignPS  & 90.3    & 81.2 &\textbf{93.1} & \textbf{93.4}\\\hline
\end{tabular}
\caption{Comparative results on CUHK-SYSU with different training structures. }
\label{tab:tasks}
\end{table}

\begin{table}[t]
\setlength{\abovecaptionskip}{1mm}
\centering
\begin{tabular}{p{2.4cm}|p{0.8cm}<{\centering}p{0.8cm}<{\centering}|p{1.1cm}<{\centering}p{1.2cm}<{\centering}}
\hline\thickhline
\rowcolor{mygray} 
{Methods}  & mAP & top-1  & $\Delta$ mAP  & $\Delta$ top-1  \\  \hline \hline     
OIM
   & 92.4   & 92.9       & -   & -  \\
TOIM w/o LUT  & 92.8   & 93.2       & +0.4   & +0.3   \\ 
TOIM  w/ LUT  &\textbf{93.1} & \textbf{93.4} & +0.7    & +0.5\\\hline
\end{tabular}
\caption{Comparative results on CUHK-SYSU with different loss functions. }
\label{tab:toim}
\end{table}

\textbf{TOIM Loss}.
We evaluate the performance of our framework when adopting different loss functions and report the results in Table~\ref{tab:toim}. We find that directly employing a triplet loss brings slight improvement. When employing the items in the LUT, the TOIM improves the mAP and top-1 accuracy by 0.7\% and 0.5\%, respectively. This indicates that it is beneficial to consider the relations between the input features and the features stored in the LUT.

\begin{table}[t]
\setlength{\abovecaptionskip}{1mm}
\centering
\begin{tabular}{p{1.7cm}|p{2.7cm}|p{1.1cm}<{\centering}p{1.1cm}<{\centering}}
\hline\thickhline
\rowcolor{mygray} 
{Backbones}  & Deformable conv   &  mAP  &  top-1  \\  \hline \hline     
ResNet-50
   & none       & 93.1   & 93.4  \\
ResNet-50
   & res3      & 93.5   & 93.9  \\
ResNet-50
   & res3{\,}\&{\,}res4      & 93.5   & 94.0  \\
ResNet-50
   & res3{\,}\&{\,}res4{\,}\&{\,}res5     &\textbf{94.0} & \textbf{94.5} \\
\hline
\end{tabular}
\caption{Comparative results on CUHK-SYSU with different deformable conv layers in the backbone model. }
\label{tab:dcn}
\end{table}

\textbf{Deformable Conv in the Backbone.}
As shown in Table~\ref{tab:dcn}, inserting deformable convolutions into the backbone network has positive effects on our framework. However, the contribution of the deformable conv layers in the backbone network is less significant than the deformable conv layers in our AFA module, \eg, only $\sim$1\% improvement is observed with all the res3{\,}\&{\,}res4{\,}\&{\,}res5 deformable conv layers. These results indicate that the proposed AFA works as the key module for successful feature alignment.

\begin{table}[t]
\setlength{\abovecaptionskip}{1mm}
\centering
\begin{tabular}{p{0.3cm}|p{2.3cm}|p{0.8cm}<{\centering}p{0.8cm}<{\centering}|p{0.8cm}<{\centering}p{0.8cm}<{\centering}}
\hline\thickhline
\rowcolor{mygray} 
\multicolumn{2}{c|}{}  & \multicolumn{2}{c|}{CUHK-SYSU}                       & \multicolumn{2}{c}{PRW}                             \\ \cline{3-6} 
\rowcolor{mygray} 
\multicolumn{2}{c|}{\multirow{-2}{*}{Methods}}                       & \multicolumn{1}{c}{mAP} & \multicolumn{1}{c|}{top-1} & \multicolumn{1}{c}{mAP} & \multicolumn{1}{c}{top-1} \\ \hline\hline
\multirow{9}{*}{ \rotatebox{90}{one-step}}               & OIM~\cite{DBLP:conf/cvpr/XiaoLWLW17}      & 75.5  & 78.7      & 21.3   & 49.4    \\ 
 & IAN~\cite{DBLP:journals/pr/XiaoXTHWF19}    & 76.3  & 80.1 & 23.0   & 61.9 \\ 
 & NPSM~\cite{DBLP:conf/iccv/LiuFJKZQJY17}        & 77.9  & 81.2 & 24.2   & 53.1 \\
 & RCAA~\cite{DBLP:conf/eccv/ChangHSLYH18}        & 79.3  & 81.3 & -   & - \\
 & CTXG~\cite{DBLP:conf/cvpr/YanZNZXY19} & 84.1  & 86.5 & 33.4   & 73.6 \\
 & QEEPS~\cite{DBLP:conf/cvpr/MunjalATG19} & 88.9  & 89.1 & 37.1   & 76.7 \\
 & BINet~\cite{DBLP:conf/cvpr/DongZST20a}        & 90.0  & 90.7 & 45.3   & 81.7 \\
 & NAE~\cite{DBLP:conf/cvpr/ChenZYS20}        & 91.5 & 92.4 & 43.3   & 80.9 \\
 & NAE+~\cite{DBLP:conf/cvpr/ChenZYS20}        & 92.1 & 92.9 & 44.0   & 81.1 \\
 & \textbf{AlignPS} &93.1 & 93.4 &45.9 & 81.9 \\
 & \textbf{AlignPS+} &\textbf{94.0} & \textbf{94.5} &\textbf{46.1} & \textbf{82.1} \\
 \hline \hline
 \multirow{6}{*}{ \rotatebox{90}{two-step}} 
 & DPM+IDE~\cite{DBLP:conf/cvpr/ZhengZSCYT17}        & -  & - & 20.5   & 48.3 \\
 & CNN+MGTS~\cite{DBLP:conf/eccv/ChenZOYT18}        & 83.0  & 83.7 & 32.6   & 72.1 \\
 & CNN+CLSA~\cite{DBLP:conf/eccv/LanZG18}        & 87.2  & 88.5 & 38.7   & 65.0 \\
 & FPN+RDLR~\cite{DBLP:conf/iccv/HanYZTZGS19}        & 93.0 & 94.2 & 42.9  & 70.2 \\
  & IGPN~\cite{DBLP:conf/cvpr/DongZST20}        & 90.3  & 91.4 & \textbf{47.2}   & 87.0 \\
  & TCTS~\cite{DBLP:conf/cvpr/WangMCSC20}       & \textbf{93.9}  & \textbf{95.1} & 46.8   & \textbf{87.5} \\
  \hline
\end{tabular}
\caption{Comparison with the state-of-the-arts. The upper block lists the results of one-step models, while the lower block shows the results of two-step methods.}
\label{tab:sota}
\end{table}

\begin{figure*}[ht]
\setlength{\abovecaptionskip}{1mm}
\begin{center}
\includegraphics[width=\linewidth]{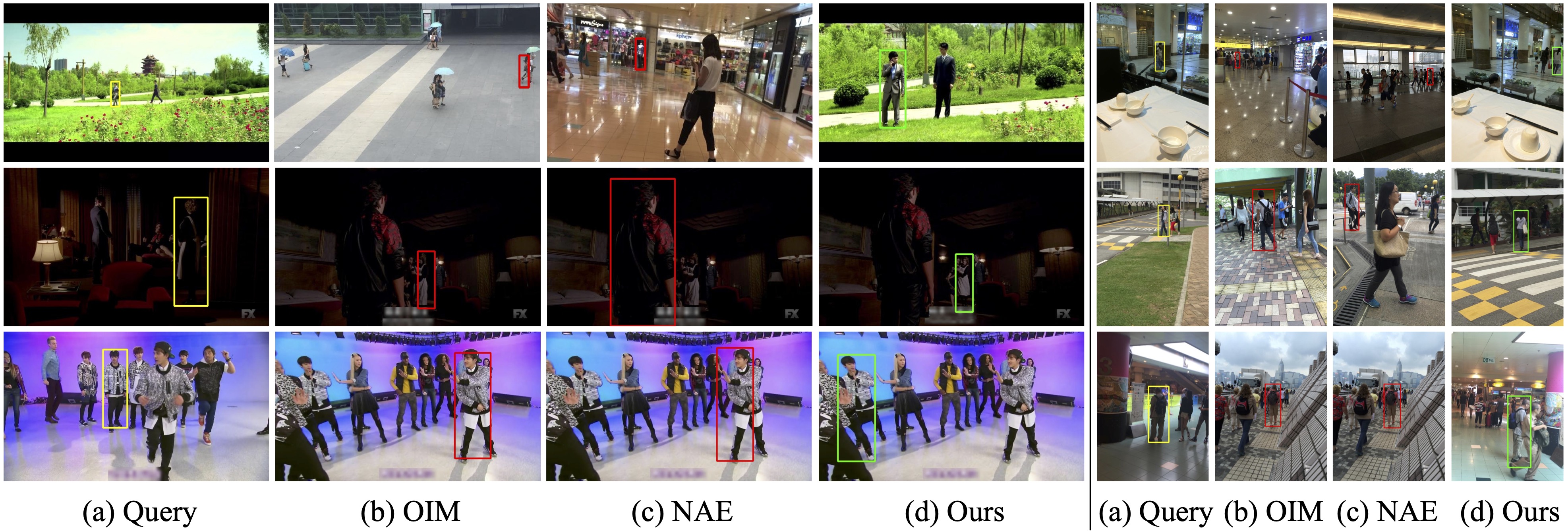}
\end{center}
\vspace{-4mm}
   \caption{Difficult cases that can be successfully retrieved by AlignPS but not OIM \cite{DBLP:conf/cvpr/XiaoLWLW17} and NAE  \cite{DBLP:conf/cvpr/ChenZYS20}.  The yellow bounding boxes denote the queries, while the green and red bounding boxes denote correct and incorrect top-1 matches, respectively.}
\label{fig:vis}
\vspace{-4mm}
\end{figure*}

\begin{figure}[t]
\vspace{-4mm}
\setlength{\abovecaptionskip}{1mm}
\centering
\subfloat[Comparison to one-step models\label{subfig-3-1}]{%
   \includegraphics[width=0.485\linewidth]{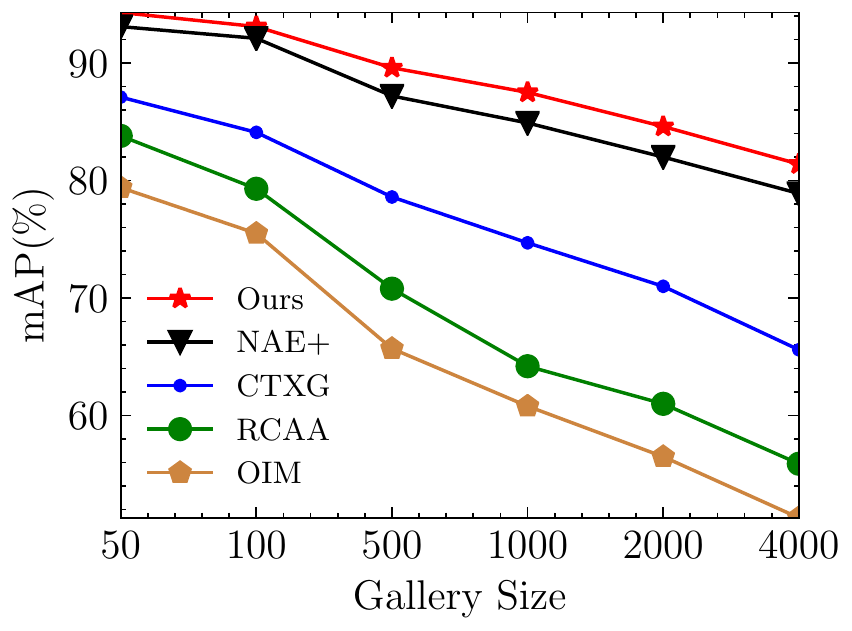}
}
 \hfill
\subfloat[Comparison to two-step models\label{subfig-3-2}]{%
   \includegraphics[width=0.485\linewidth]{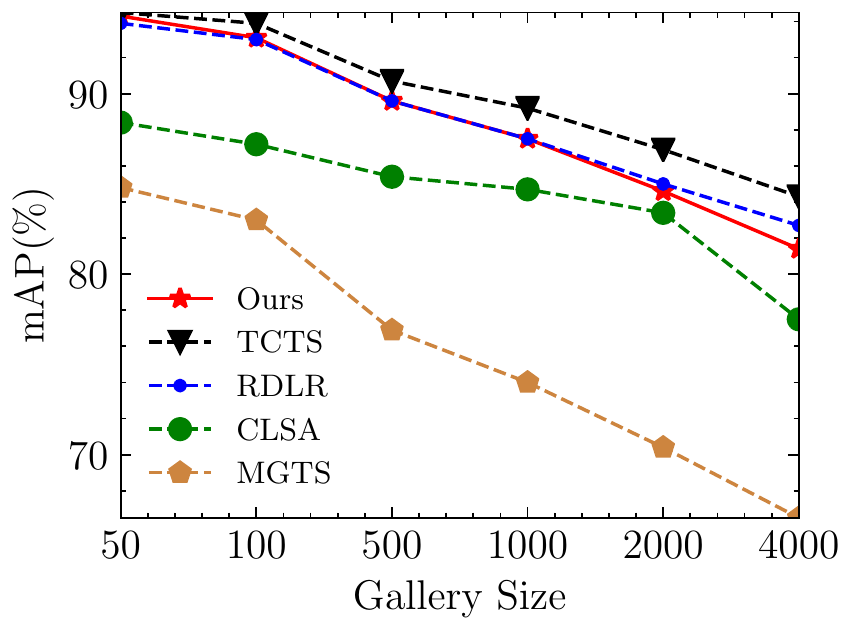}
}
 \caption{Comparative results on CUHK-SYSU with different gallery sizes. Our model (AlignPS) is compared with both (a) one-step models and (b) two-step models.}
 \label{fig:gallerysize}
\vspace{-2mm}
\end{figure}

\subsection{Comparison to the State-of-the-Arts}
We compare our model with the state-of-the-arts, including both one-step models \cite{DBLP:conf/cvpr/XiaoLWLW17,DBLP:journals/pr/XiaoXTHWF19,DBLP:conf/iccv/LiuFJKZQJY17,DBLP:conf/eccv/ChangHSLYH18,DBLP:conf/cvpr/YanZNZXY19,DBLP:conf/cvpr/MunjalATG19,DBLP:conf/cvpr/DongZST20a,DBLP:conf/cvpr/ChenZYS20} and two-step models~\cite{DBLP:conf/eccv/ChenZOYT18,DBLP:conf/eccv/LanZG18,DBLP:conf/iccv/HanYZTZGS19,DBLP:conf/cvpr/DongZST20,DBLP:conf/cvpr/WangMCSC20}. 
We denote our model with deformable conv layers in the backbone as \textbf{AlignPS+}.

\textbf{Results on CUHK-SYSU}.
As shown in Table~\ref{tab:sota}, AlignPS/AlignPS+ outperforms all one-step person search models employing two-stage detection frameworks, which require region proposals and ROI-Align for inference. In contrast, our model is anchor-free and allows single-stage inference with a very simple structure, whilst running at a higher speed. Notably, AlignPS+ outperforms the current best-performing NAE+~\cite{DBLP:conf/cvpr/ChenZYS20} by 1.9\% and 1.6\% in mAP and top-1 accuracy, respectively. Also note that our model outperforms most two-step models, despite the fact that they employ two separate models for detection and re-id.

We visualize the results of AlignPS w.r.t. mAP with various gallery sizes and compare our model with both one-step and two-step models. Fig.~\ref{fig:gallerysize} illustrates the detailed comparison results. As we can see, AlignPS outperforms all the one-step models by notable margins, and is only inferior to the strongest two-step model TCTS~\cite{DBLP:conf/cvpr/WangMCSC20}, which requires an explicitly trained re-id model to adapt to the detection results. In contrast, our model does not need such a two-step process, as the alignment between the two subtasks is performed implicitly within the framework.

\textbf{Results on PRW}.
PRW contains less training data; therefore, all the models achieve worse performance on this dataset. Nevertheless, as can be observed from Table~\ref{tab:sota}, our model still outperforms all the one-step methods. We notice that BINet~\cite{DBLP:conf/cvpr/DongZST20a} also achieves strong performance on PRW. However, it requires an additional re-id branch to achieve region alignment during training, while our model efficiently addresses this issue with the AFA module.

\textbf{Efficiency Comparison}.
Since different methods are evaluated with different GPUs, it is difficult to conduct a fair comparison of the efficiency of all the models. Here, we compare our method with OIM\footnote{We test the PyTorch implementation at \url{https://github.com/serend1p1ty/person_search}} \cite{DBLP:conf/cvpr/XiaoLWLW17} and NAE/NAE+~\cite{DBLP:conf/cvpr/ChenZYS20} on the same Tesla V100 GPU. All the test images are resized to 1500$\times$900 before being fed to the networks. As shown in Table~\ref{tab:runtime}, our anchor-free AlignPS only takes 61 milliseconds to process an image, which is 27\% and 38\% faster than NAE and NAE+, respectively. 
For query-guided models, \eg, IGPN~\cite{DBLP:conf/cvpr/DongZST20} and QEEPS~\cite{DBLP:conf/cvpr/MunjalATG19}, they needs to re-compute all the gallery features given each query. As AlignPS only computes the gallery features once, the total computation of these models can be thousands of times of AlignPS. It is also noteworthy that the parameters of all the two-step models are twice as our framework. These results clearly demonstrate the advantage of our anchor-free model in terms of computational efficiency.

\begin{table}[t]
\setlength{\abovecaptionskip}{1mm}
\centering
\begin{tabular}{p{1.8cm}p{3cm}p{0.6cm}p{1.3cm}<{\centering}}
\hline\thickhline
\rowcolor{mygray} 
Methods &Backbones & GPU  &  Time (ms)  \\  \hline \hline     
OIM \cite{DBLP:conf/cvpr/XiaoLWLW17} & ResNet-50 & V100 & 118 \\
NAE+ \cite{DBLP:conf/cvpr/ChenZYS20} & ResNet-50 & V100 & 98 \\
NAE \cite{DBLP:conf/cvpr/ChenZYS20} & ResNet-50 & V100 & 83 \\
\textbf{AlignPS} & ResNet-50 & V100 & \textbf{61} \\
\textbf{AlignPS+} & ResNet-50 w/ dconv & V100 & 67
\\\hline
\end{tabular}
\caption{Runtime comparison of different models.}
\label{tab:runtime}
\vspace{-4mm}
\end{table}

\textbf{Qualitative Results}.
Some qualitative results are illustrated in Fig.~\ref{fig:vis}, where the query images come from movies/TV (left) and hand-held cameras (right). We can observe that our model successfully handles occlusions and scale/viewpoint variations, where OIM \cite{DBLP:conf/cvpr/XiaoLWLW17} and NAE \cite{DBLP:conf/cvpr/ChenZYS20} fail, demonstrating the robustness of our AlignPS.

\section{Conclusion}
In this paper, we propose the first anchor-free model to simplify the framework for person search, where detection and re-id are jointly addressed by a one-step model. We also design the aligned feature aggregation module to effectively address the scale, region, and task misalignment issues when accommodating an anchor-free detector for the person search task. Extensive experiments demonstrate that the proposed framework not only outperforms existing person search methods, but also runs at a higher speed.

{\small
\bibliographystyle{ieee_fullname}
\bibliography{egbib}
}

\end{document}